\documentclass[11pt]{article}

\usepackage[margin=1in]{geometry}
\usepackage{lmodern}
\usepackage[hyphens,spaces]{url}
\usepackage{hyperref}
\usepackage{tabularx}
\usepackage{booktabs}
\usepackage{xcolor}
\usepackage{caption}
\usepackage{enumitem}
\usepackage{amssymb}
\usepackage{tikz}
\usetikzlibrary{
  patterns,
  positioning,
  arrows.meta,
  shapes.geometric,
  shapes.symbols,
  fit,
  calc,
  backgrounds
}

\definecolor{ProviderBlue}{RGB}{170,220,255}
\definecolor{APIBlue}{RGB}{50,110,255}
\definecolor{ToolsPurple}{RGB}{190,175,255}
\definecolor{ToolYellow}{RGB}{245,235,170}
\definecolor{ServerGreen}{RGB}{40,160,70}
\definecolor{ClientOrange}{RGB}{245,166,35}

\title{OGX: An Open-Source, Vendor-Neutral Generative AI Application Server}

\author{}
\date{June 2026}

\makeatletter
\renewcommand{\maketitle}{%
  \begin{center}
    {\LARGE\bfseries \@title\footnote{OGX (Open GenAI Stack): \url{https://github.com/ogx-ai/ogx} \,|\, Docs: \url{https://ogx-ai.github.io}} \par}
    \vspace{1em}
    {\large
      Francisco Javier Arceo\textsuperscript{1},
      S\'ebastien Han\textsuperscript{1},
      Matthew Farrellee\textsuperscript{3},
      Charlie Doern\textsuperscript{1},
      Yuan Tang\textsuperscript{1},
      Derek Higgins\textsuperscript{1},
      Varsha Prasad Narsing\textsuperscript{1},
      Gordon Sim\textsuperscript{1},
      Sumanth Kamenani\textsuperscript{1},
      Ben Browning\textsuperscript{1},
      Raghotham Murthy\textsuperscript{2}
      \par}
    \vspace{0.5em}
    {\normalsize \textsuperscript{1}Red Hat AI \quad \textsuperscript{2}Meta \quad \textsuperscript{3}Independent \par}
    \vspace{0.5em}
    {\normalsize \@date \par}
  \end{center}
  \vspace{1.5em}
}
\makeatother

\begin{document}
\maketitle

\begin{abstract}
OGX (Open GenAI Stack) is an open-source AI application server and Python library that implements the APIs of major frontier labs (OpenAI, Anthropic, Google) with pluggable backend providers. Developers building agentic AI applications---such as retrieval-augmented generation pipelines, multi-turn agents, and tool-calling workflows---can develop against a single API surface and deploy with any combination of inference engine, vector database, and safety backend, without changing application code. OGX's primary focus is the Responses API for server-side agentic orchestration, conforming to the Open Responses specification. The server also supports the Anthropic Messages API and Google GenAI Interactions API, decoupling SDK choice from model and deployment decisions. With over 20 inference providers, 13 vector store backends, and a companion Kubernetes Operator for production deployment, OGX serves as the self-hosted, model-agnostic backend for AI-powered developer tools including Claude Code, Codex CLI, OpenCode, and OpenHands. The project has over 8,400 GitHub stars, 242 contributors, and 4,000 commits across nearly two years of public development.
\end{abstract}

\section{Introduction}

AI application development today is tightly coupled to proprietary API providers. While inference-only workloads can increasingly be swapped across providers---vLLM, for example, supports the Responses API for basic inference---applications that rely on the full stack (retrieval, tool calling, conversation state, safety guardrails) remain difficult to migrate without rewriting significant application logic. This coupling limits deployment flexibility and prevents organizations from running AI workloads on their own infrastructure---a requirement in regulated industries and air-gapped environments.

The problem is compounded by the emergence of AI-powered developer tools. Tools like Claude Code, Codex CLI, OpenCode, and OpenHands need a backend that can serve models, execute tools, manage conversation state, and handle retrieval---all through standard APIs. Organizations that want to self-host these capabilities currently must assemble and integrate multiple systems: an inference engine, a vector database, a tool execution runtime, and custom glue code to connect them.

OGX addresses this by providing a complete, self-hosted AI application server that implements the OpenAI API surface---with the Responses API as its primary focus---alongside Anthropic Messages and Google GenAI Interactions compatibility layers. It decouples three decisions that are currently entangled: which SDK to use, which model to run, and where to deploy. Developers write code against standard endpoints and swap the underlying infrastructure through configuration, not code changes.

This paper describes OGX's design, architecture, and positioning in the AI application ecosystem. Section~\ref{sec:field} surveys the state of the field. Section~\ref{sec:design} details the software design. Section~\ref{sec:operator} describes the Kubernetes Operator. Section~\ref{sec:impact} discusses research impact and adoption.

\section{State of the Field}
\label{sec:field}

The AI application ecosystem has stratified into layers that each solve a subset of the deployment problem. Table~\ref{tab:comparison} summarizes the landscape.

\textbf{Inference engines} (vLLM~\cite{kwon2023vllm}, SGLang~\cite{sglang}, Ollama) focus on efficient model serving. They optimize throughput and latency but do not provide retrieval, conversation state, tool execution, or safety guardrails. An application using vLLM for inference must separately integrate a vector database, implement its own agentic loop, and manage multi-turn state.

\textbf{API gateways} (LiteLLM, OpenRouter) provide a unified interface across multiple inference providers but act as pass-through proxies. They do not manage vector stores, execute tool calls, or maintain conversation history---they translate request formats between SDKs and providers.

\textbf{Client-side frameworks} (LangChain~\cite{langchain}, LangGraph~\cite{langgraph}, LlamaIndex~\cite{llamaindex}, CrewAI~\cite{crewai}, Haystack~\cite{haystack}) provide rich developer abstractions for building agents and RAG pipelines. However, they execute orchestration client-side, distributing security-critical logic across application code. These frameworks are complementary to OGX: they compose agent logic while OGX provides the server-side execution target they call into.

\textbf{Proprietary platforms} (OpenAI's Responses API~\cite{openaiResponsesAPI}, Databricks Mosaic AI~\cite{databricksAgentFramework}) offer integrated experiences but couple applications to a specific vendor's infrastructure and pricing.

OGX occupies a distinct position: a self-hosted, vendor-neutral server that implements the full API surface---inference, retrieval, tool execution, conversation management, and safety---with pluggable providers at every layer. Its conformance to the Open Responses specification~\cite{openresponses} ensures interoperability with any client that speaks the same protocol.

\begin{table}[htbp]
  \centering
  \small
  \setlength{\tabcolsep}{4pt}
  \renewcommand{\arraystretch}{1.15}
  \begin{tabularx}{\textwidth}{@{}l|ccccc>{\raggedright\arraybackslash}X@{}}
  \toprule
  System & Inference & RAG & Tools & State & Safety & Deployment \\
  \midrule
  vLLM / SGLang & $\checkmark$ & -- & -- & -- & -- & Self-hosted \\
  Ollama & $\checkmark$ & -- & -- & -- & -- & Local \\
  LiteLLM & $\checkmark$$^*$ & -- & -- & -- & -- & Gateway \\
  LangChain / LangGraph & $\checkmark$$^*$ & $\checkmark$ & $\checkmark$ & $\checkmark$ & -- & Client-side \\
  OpenAI Platform & $\checkmark$ & $\checkmark$ & $\checkmark$ & $\checkmark$ & $\checkmark$ & SaaS \\
  \textbf{OGX} & $\checkmark$ & $\checkmark$ & $\checkmark$ & $\checkmark$ & $\checkmark$ & Local \& Self-hosted \\
  \bottomrule
  \end{tabularx}
  \caption{Feature comparison across AI application system categories. $^*$Proxy/wrapper---delegates to external provider.}
  \label{tab:comparison}
\end{table}

\section{Software Design}
\label{sec:design}

\subsection{Provider Architecture}

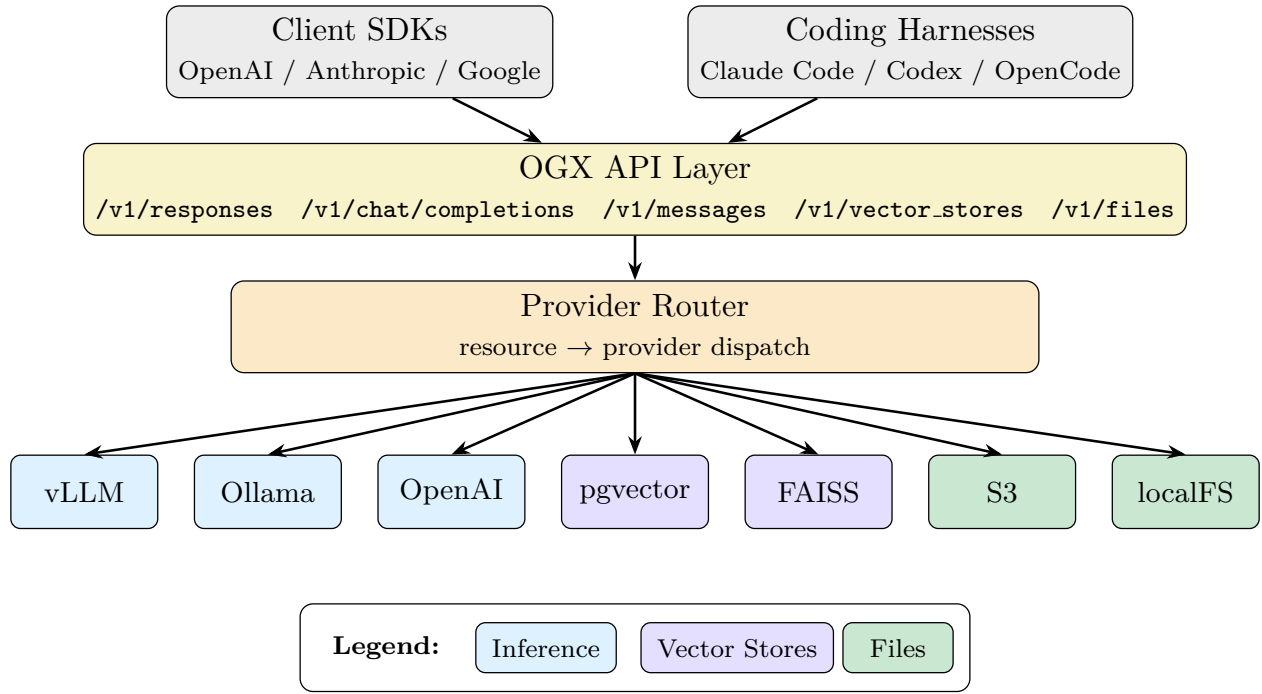
\begin{figure}[htbp]
  \centering
  \resizebox{\textwidth}{!}{%
  \begin{tikzpicture}[
    box/.style={draw, rounded corners=4pt, minimum height=10mm,
                align=center, font=\small},
    provider/.style={draw, rounded corners=3pt, minimum width=16mm, minimum height=8mm,
                     align=center, font=\footnotesize, fill=ProviderBlue!40},
    arrow/.style={-{Stealth[length=2mm]}, thick},
  ]
    \node[box, fill=gray!15, minimum width=42mm] (client) at (-3,4) {Client SDKs\\{\scriptsize OpenAI / Anthropic / Google}};
    \node[box, fill=gray!15, minimum width=42mm] (harness) at (3,4) {Coding Harnesses\\{\scriptsize Claude Code / Codex / OpenCode}};

    \node[box, fill=ToolYellow!60, minimum width=88mm] (api) at (0,2.5)
      {OGX API Layer\\{\scriptsize\texttt{/v1/responses} ~ \texttt{/v1/chat/completions} ~ \texttt{/v1/messages} ~ \texttt{/v1/vector\_stores} ~ \texttt{/v1/files}}};

    \node[box, fill=ClientOrange!25, minimum width=88mm] (router) at (0,1)
      {Provider Router\\{\scriptsize resource $\rightarrow$ provider dispatch}};

    \node[provider] (p1) at (-6,-0.8) {vLLM};
    \node[provider] (p2) at (-4,-0.8) {Ollama};
    \node[provider] (p3) at (-2,-0.8) {OpenAI};
    \node[provider, fill=ToolsPurple!40] (p4) at (0,-0.8) {pgvector};
    \node[provider, fill=ToolsPurple!40] (p5) at (2,-0.8) {FAISS};
    \node[provider, fill=ServerGreen!25] (p6) at (4,-0.8) {S3};
    \node[provider, fill=ServerGreen!25] (p7) at (6,-0.8) {localFS};

    \draw[arrow] (client) -- (api);
    \draw[arrow] (harness) -- (api);
    \draw[arrow] (api) -- (router);
    \draw[arrow] (router.south) -- (p1.north);
    \draw[arrow] (router.south) -- (p2.north);
    \draw[arrow] (router.south) -- (p3.north);
    \draw[arrow] (router.south) -- (p4.north);
    \draw[arrow] (router.south) -- (p5.north);
    \draw[arrow] (router.south) -- (p6.north);
    \draw[arrow] (router.south) -- (p7.north);

    \node[draw, rounded corners=4pt, inner sep=5pt, fill=white] (legend) at (0,-2.5) {%
      \begin{tikzpicture}[baseline=0pt]
        \node[font=\scriptsize\bfseries, anchor=east] at (0,0) {Legend:};
        \node[provider, minimum width=14mm, minimum height=5mm, font=\scriptsize, anchor=west] at (0.2,0) {Inference};
        \node[provider, fill=ToolsPurple!40, minimum width=18mm, minimum height=5mm, font=\scriptsize, anchor=west] at (2.0,0) {Vector Stores};
        \node[provider, fill=ServerGreen!25, minimum width=12mm, minimum height=5mm, font=\scriptsize, anchor=west] at (4.2,0) {Files};
      \end{tikzpicture}%
    };
  \end{tikzpicture}%
  }
  \caption{Provider architecture: client requests are dispatched through the API layer and router to pluggable backend providers. Representative APIs and providers shown; OGX additionally exposes Containers, Skills, Safety, Prompts, Conversations, and Telemetry APIs.}
  \label{fig:provider-arch}
\end{figure}

OGX's core abstraction is the pluggable provider. Each API capability---inference, vector storage, safety, tool runtime, file processing---is defined by a Protocol interface. Concrete providers implement these interfaces for specific backends (Figure~\ref{fig:provider-arch}):
\begin{itemize}[nosep]
  \item \textbf{Inference:} \texttt{remote::openai}, \texttt{remote::anthropic}, \texttt{remote::vllm}, \texttt{remote::ollama}, \texttt{remote::bedrock}, \texttt{remote::ge\-mi\-ni}, and 15+ others.
  \item \textbf{Vector stores:} \texttt{inline::faiss}, \texttt{inline::sqlite-vec}, \texttt{remote::pgvector}, \texttt{remote::qdrant}, \texttt{remote::milvus}, \texttt{remote::weaviate}, \texttt{remote::chromadb}, and others.
  \item \textbf{Safety:} Content moderation providers for input/output guardrails.
  \item \textbf{Tools:} Built-in tools (file search, web search, code interpreter) and MCP~\cite{mcp} integration for external tool servers.
\end{itemize}

A routing layer dispatches requests to provider instances based on logical resource identifiers, enabling multiple providers to serve the same API simultaneously. For example, Ollama can handle local models while OpenAI handles hosted models, both behind \texttt{/v1/chat/completions}. The routing table tracks which provider owns each resource (model, vector store, file), enabling auto-dispatch without client-side logic.

\subsection{Distributions}

A \emph{distribution} packages a specific set of providers and configuration into a deployable unit, decoupling application logic from infrastructure selection. Developers prototype with lightweight inline providers (Ollama for inference, sqlite-vec~\cite{sqlitevec} for vectors) and deploy to production backends (vLLM~\cite{kwon2023vllm}, pgvector) by changing the distribution configuration, not the application code. OGX ships a \texttt{starter} distribution for quick setup and supports custom distributions for production environments.

\subsection{Dual Deployment Model}

OGX runs in two modes:
\begin{itemize}[nosep]
  \item \textbf{Server mode:} HTTP endpoints accessible from any language or tool. This is the production deployment model.
  \item \textbf{Library mode:} Direct Python import with zero network overhead. Suitable for notebooks, scripts, and rapid prototyping.
\end{itemize}
Both modes use identical provider routing and API semantics. The recommended progression is: start with the library for prototyping, graduate to the server when you need multi-language access, scaling, or production isolation.

\subsection{Multi-SDK Compatibility}

OGX serves three client SDK protocols from a single server instance:
\begin{itemize}[nosep]
  \item \textbf{OpenAI-compatible endpoints} (\texttt{/v1/chat/completions}, \texttt{/v1/responses}, \texttt{/v1/vector\_stores}): the primary interface. The Responses API implementation conforms to the Open Responses specification~\cite{openresponses}.
  \item \textbf{Anthropic Messages endpoint} (\texttt{/v1/messages}): native compatibility for teams using the Anthropic SDK.
  \item \textbf{Google GenAI Interactions endpoint} (\texttt{/v1alpha/interactions}): native compatibility for teams using the Google GenAI SDK.
\end{itemize}

This means SDK choice, model selection, and deployment target are fully independent decisions. A team can use the Anthropic SDK to call an open-weight model served by vLLM through OGX, or use the OpenAI SDK to call Anthropic's Claude---the server handles the translation.

\subsection{Server-Side Agentic Orchestration}

The Responses API~\cite{openaiResponsesAPI} is OGX's primary API focus and implements server-side agentic orchestration: the inference$\rightarrow$tool$\rightarrow$inference loop executes within the server process, not the client. This design centralizes several critical concerns:

\begin{itemize}[nosep]
  \item \textbf{Tool authorization:} Tool calls are executed server-side with centralized authorization, not delegated to potentially untrusted clients.
  \item \textbf{Conversation state:} Multi-turn state is managed server-side with tenant-scoped isolation.
  \item \textbf{Safety guardrails:} Input and output safety checks are applied at each step of the agentic loop.
  \item \textbf{Context management:} A Compaction API summarizes long conversation histories to manage context window limits.
\end{itemize}

Built-in tools include file search (RAG over vector stores with hybrid dense/sparse retrieval), web search, code interpretation, computer use, and Model Context Protocol (MCP)~\cite{mcp} integration for external tool servers.

For multitenant deployments, OGX provides attribute-based access control (ABAC) that enforces tenant isolation at the retrieval, tool execution, state management, and API routing layers. The security properties of this architecture---specifically, how ABAC-gated retrieval eliminates cross-tenant data leakage while adding negligible overhead---have been formally analyzed and empirically validated~\cite{arceo2026securing}.

\subsection{API Surface}

OGX exposes over 20 APIs covering the full lifecycle of AI applications (Table~\ref{tab:apis}):

\begin{itemize}[nosep]
  \item \textbf{Inference APIs:} Chat Completions and Embeddings provide standard inference endpoints. The Responses API serves as the central agentic orchestration endpoint, coordinating multi-turn conversations, tool calling, and server-side execution~\cite{openaiResponsesAPI}.
  \item \textbf{Data APIs:} Vector Stores and Search APIs support dense, sparse, and hybrid retrieval with structured metadata filtering for tenant isolation. The Files and File Processors APIs provide tenant-scoped file storage (S3, GCS, PVCs, local filesystem) with parsing and chunking for vector store ingestion. The Batches API supports offline processing.
  \item \textbf{Agent and Tool APIs:} Built-in tools include file search (RAG over vector stores), web search, code interpreter, shell (via Containers), image generation, and computer use. External tools integrate via MCP~\cite{mcp}. Function Calling and Connectors enable custom tool integration.
  \item \textbf{Execution APIs:} The Containers API (\texttt{/v1/containers}) manages sandboxed execution environments---matching the OpenAI Containers API---enabling models to execute shell commands in isolated Docker, Podman, or Kubernetes containers via a \texttt{shell} tool in the Responses API. The Skills API (\texttt{/v1alpha/skills}) manages versioned skill bundles that package tools, prompts, and configuration into reusable, composable units for domain-specific capabilities.
  \item \textbf{State APIs:} The Conversations API persists multi-turn history with tenant-scoped isolation, eliminating client-side session stores. The Prompts API provides versioned prompt template management. A resource registry tracks models, vector stores, files, and tool groups as first-class server objects with ownership metadata. The Compaction API automatically summarizes long conversation histories to manage context window limits. This server-side state layer is a distinguishing feature---most inference engines and API gateways treat requests as stateless, requiring applications to build their own persistence. OGX manages this natively, enabling it to function as a complete application server rather than a stateless proxy.
  \item \textbf{Safety APIs:} Content moderation providers apply input and output guardrails at each step of the agentic loop.
  \item \textbf{Admin APIs:} Model registry for managing available models across providers. Telemetry via OpenTelemetry (OTEL) for observability and tracing of agent execution, with built-in MLflow~\cite{mlflow} tracing integration (which supports OTEL) for logging spans, tool calls, and retrieval steps to existing ML experiment tracking infrastructure.
\end{itemize}

\begin{table}[htbp]
  \centering
  \small
  \setlength{\tabcolsep}{4pt}
  \renewcommand{\arraystretch}{1.15}
  \begin{tabularx}{\textwidth}{@{}l|X@{}}
  \toprule
  Category & APIs \\
  \midrule
  Inference & Chat Completions, Responses (agentic orchestration), Embeddings \\
  Data & Vector Stores, Search, Files, File Processors, Batches \\
  Agent/Tools & File search, web search, code interpreter, shell, image generation, computer use, MCP, Function Calling, Connectors \\
  Execution & Containers (sandboxed environments), Skills (versioned bundles) \\
  State & Conversations, Prompts, Compaction, Resource Registry \\
  Safety & Content moderation (input/output guardrails) \\
  Admin & Models, Telemetry (OTEL, MLflow tracing) \\
  \bottomrule
  \end{tabularx}
  \caption{OGX API surface organized by category.}
  \label{tab:apis}
\end{table}

\section{Example Usage}
\label{sec:examples}

The following examples demonstrate OGX's portability: the same client code works regardless of which inference provider or vector store backend is configured on the server.

\subsection{RAG Agent with the OpenAI SDK}

Building a RAG agent requires only standard OpenAI SDK calls. The server handles document chunking, embedding, vector storage, retrieval, and context injection transparently:

\begin{verbatim}
from openai import OpenAI

client = OpenAI(base_url="http://localhost:8321/v1",
                api_key="unused")

# Create a vector store and upload documents
vector_store = client.vector_stores.create(name="docs")
client.vector_stores.files.upload(
    vector_store_id=vector_store.id,
    file=open("manual.pdf", "rb"),
)

# Query with server-side RAG via the Responses API
response = client.responses.create(
    model="meta-llama/Llama-3.2-3B-Instruct",
    input="What are the installation requirements?",
    tools=[{"type": "file_search",
            "vector_store_ids": [vector_store.id]}],
)
print(response.output_text)
\end{verbatim}

Switching from a local Ollama backend to a production vLLM cluster requires changing only the server's distribution configuration---the client code above remains identical.

\subsection{Published Use Cases}

OGX's provider portability has been demonstrated across diverse enterprise backends:

\begin{itemize}[nosep]
  \item \textbf{IBM watsonx.ai + Milvus:} Enterprise RAG pipeline using watsonx.ai for inference and watsonx.data Milvus for vector storage, with Llama Stack as the unifying orchestration layer~\cite{ibm_rag_milvus}.
  \item \textbf{Oracle Cloud Infrastructure:} Generative AI application development using OCI AI Blueprints with Llama Stack for standardized API access~\cite{oracle_oci_ogx}.
  \item \textbf{Red Hat OpenShift:} An intelligent operations agent combining agentic RAG, web search, and MCP tool integration (OpenShift cluster management, Slack notifications) for automated incident response~\cite{redhat_ops_agent}.
\end{itemize}

The framework has been presented at Meta Connect~\cite{meta_connect_ogx} and IBM TechXchange~\cite{ibm_techxchange_ogx} as a standardization layer for enterprise AI applications.

\section{Kubernetes Operator}
\label{sec:operator}

The OGX Kubernetes Operator~\cite{ogxk8soperator} provides declarative, production-grade deployment through the \texttt{OGXServer} custom resource definition (CRD). Written in Go using the operator-sdk framework, it automates the full lifecycle of OGX server deployments on Kubernetes and OpenShift.

\subsection{Custom Resource Model}

A single \texttt{OGXServer} CR specifies:
\begin{itemize}[nosep]
  \item \textbf{Distribution:} Which AI stack variant to deploy (e.g., \texttt{starter}, custom distributions).
  \item \textbf{Workload:} Replica count, persistent storage size and mount paths, environment variable overrides for inference model configuration.
  \item \textbf{Network:} External access configuration (hostname-based routing) and network policies (enabled by default per-CR).
\end{itemize}

The operator reconciles the desired state expressed in the CR with the actual cluster state, handling creation, scaling, updates, and teardown of OGX server pods.

\subsection{Operational Features}

\begin{itemize}[nosep]
  \item \textbf{ConfigMap-driven image overrides:} Administrators update the \texttt{ogx-operator-config} ConfigMap with an \texttt{image-overrides} key, and all matching \texttt{OGXServer} resources restart with the new image---enabling fleet-wide image updates without redeploying the operator.
  \item \textbf{ConfigMap-based configuration:} Users supply \texttt{config.yaml} content via ConfigMaps; the operator watches for changes and automatically restarts pods to load updated configuration.
  \item \textbf{Multi-architecture builds:} Supports \texttt{linux/amd64} and \texttt{linux/arm64} with FIPS-compliant images built using native architecture-matched CI runners.
  \item \textbf{Dual platform support:} First-class support for both vanilla Kubernetes (using cert-manager for webhook TLS) and OpenShift (using built-in service-serving-cert-signer).
  \item \textbf{Quickstart scripts:} \texttt{hack/deploy-quickstart.sh} enables rapid setup with provider and model flags.
\end{itemize}

\subsection{Isolation Topologies}

The operator supports three deployment topologies:
\begin{enumerate}[nosep]
  \item \textbf{Shared instances:} Multiple tenants share a single OGX server with ABAC-enforced logical isolation.
  \item \textbf{Per-tenant instances:} Namespace-level isolation with Kubernetes RBAC, each tenant receiving a dedicated OGX server.
  \item \textbf{Hybrid:} Shared inference with per-tenant data paths, balancing cost efficiency with isolation requirements.
\end{enumerate}

\section{Research Impact and Adoption}
\label{sec:impact}

\subsection{Production Deployments}

OGX is deployed in production across enterprises in telecommunications, semiconductor manufacturing, financial services, insurance, and consulting. These deployments use the full stack: pluggable inference providers, tenant-isolated vector stores, server-side agentic orchestration, and Kubernetes-based deployment via the operator. Published use cases span IBM watsonx.ai with Milvus vector storage~\cite{ibm_rag_milvus}, Oracle Cloud Infrastructure with OCI AI Blueprints~\cite{oracle_oci_ogx}, and Red Hat OpenShift with MCP-based operational agents~\cite{redhat_ops_agent}.

\subsection{Academic Validation}

The security architecture of OGX's multitenant isolation model was formally analyzed and empirically validated in a peer-reviewed publication at the ACM Conference on AI and Agentic Systems (CAIS '26)~\cite{arceo2026securing}. The evaluation demonstrated that ABAC-gated retrieval eliminates cross-tenant data leakage (0\% cross-tenant leakage rate) while adding approximately 19ms to the search path. The defense operates at the retrieval layer, making it resilient to prompt injection attacks regardless of model behavior.

\subsection{Community and Ecosystem}

As of June 2026, the project has:
\begin{itemize}[nosep]
  \item Over 8,400 GitHub stars and 1,300 forks.
  \item 242 unique contributors and over 4,000 commits.
  \item 68 releases across nearly two years of public development (since July 2024).
  \item Weekly community contributor calls and an active Discord server.
  \item Integrations contributed by external organizations including Red Hat, IBM, Oracle, and Infinispan.
\end{itemize}

OGX conforms to the Open Responses specification~\cite{openresponses} and serves as a reference implementation for open, vendor-neutral agentic AI APIs.

\subsection{AI-Powered Developer Tools}

OGX is designed to serve as the backend for AI-powered developer tools that implement OpenAI-compatible APIs, including Claude Code, Codex CLI, OpenCode, and OpenHands. By providing a self-hosted, model-agnostic server that speaks these protocols, OGX enables organizations to use these tools with any model on their own infrastructure.

\section{Acknowledgements}

We thank Meta for creating and open-sourcing Llama Stack, the foundation from which OGX evolved. We are grateful to Red Hat for supporting the development of OGX. We thank the OGX contributor community for their sustained contributions to the project.

\bibliographystyle{plain}
\bibliography{references}

\end{document}